\documentclass{article}
\PassOptionsToPackage{table}{xcolor}
\usepackage{iclr2027_conference,times}

\usepackage{amsmath,amsfonts,bm}

\def\eqref#1{equation~\ref{#1}}

\def\1{\bm{1}}

\DeclareMathAlphabet{\mathsfit}{\encodingdefault}{\sfdefault}{m}{sl}
\SetMathAlphabet{\mathsfit}{bold}{\encodingdefault}{\sfdefault}{bx}{n}

\usepackage{booktabs}
\usepackage{multirow}
\usepackage{graphicx}
\usepackage{marvosym}
\usepackage{xcolor}
\usepackage{float}
\usepackage{tabularx}

\newcolumntype{C}{>{\centering\arraybackslash}X}
\newif\ifshowrevisions
\showrevisionsfalse

\makeatletter
\newcommand{\includesvggraphic}[2][]{%
  \begingroup
  \filename@parse{#2}%
  \edef\svgpdfcompanion{\filename@area\filename@base.pdf}%
  \includegraphics[#1]{\svgpdfcompanion}%
  \endgroup
}
\makeatother
\usepackage{placeins}
\usepackage[colorlinks=true,allcolors=blue]{hyperref}

\hypersetup{
  colorlinks=true,
  linkcolor=blue,
  citecolor=blue,
  urlcolor=black,
  pdfborder={0 0 1},
  citebordercolor={0 1 0},
  linkbordercolor={1 0 0},
  urlbordercolor={0 1 1},
  pdftitle={Multimodal Flow: Unified Flow Modeling of Language and Vision in Embedding Spaces},
  pdfauthor={Hongyuan Tao; Xinggang Wang; Lianghui Zhu; Yongkang Li; Yunchao Wei; Bin Feng; Shaoyu Chen; Qian Zhang; Chang Huang; Kai Yu}
}
\usepackage{url}

\newcommand{\codeurl}{https://github.com/hustvl/Multimodal-Flow}

\title{Multimodal Flow: Unified Flow Modeling of Language and Vision in Embedding Spaces}
\author{
  \begin{tabular}{l}
  \normalfont
  {\bfseries
  Hongyuan Tao\textsuperscript{1}\hspace{0.2em}
  Xinggang Wang\textsuperscript{1},\textsuperscript{\Letter}\hspace{0.2em}
  Lianghui Zhu\textsuperscript{1}\hspace{0.2em}
  Yongkang Li\textsuperscript{1}\hspace{0.2em}
  Yunchao Wei\textsuperscript{2}\hspace{0.2em}
  Bin Feng\textsuperscript{1}
  }\\
  {\bfseries
  Shaoyu Chen\textsuperscript{3}\hspace{0.2em}
  Qian Zhang\textsuperscript{3}\hspace{0.2em}
  Chang Huang\textsuperscript{3}\hspace{0.2em}
  Kai Yu\textsuperscript{3}
  }\\[4pt]
  {\normalfont\textsuperscript{1}Huazhong University of Science and Technology\enspace
  \textsuperscript{2}Beijing Jiaotong University\enspace
  \textsuperscript{3}Horizon Robotics}
  \\
  {\normalfont\texttt{\{hongyuantao, xgwang\}@hust.edu.cn}}
  \\
  {\normalfont\textbf{Code:}\enspace
  \href{\codeurl}{\textcolor{blue}{Project repository}}}
  \end{tabular}
}

\iclrfinalcopy

\begin{document}

\maketitle
\pagestyle{plain}
\raggedbottom

\begin{abstract}

We present \textbf{Multimodal Flow}, a fully continuous generative model of language and vision.
Most unified multimodal models either model both language and quantized images as discrete tokens or combine discrete language prediction with continuous image generation.
The former introduces a visual quantization bottleneck.
The latter requires modality-dependent objectives and sampling procedures.
Fully continuous modeling avoids these trade-offs and enables a shared generative process, but remains underexplored for multimodal pretraining.
Multimodal Flow introduces a unified continuous architecture that integrates multimodal continuous representations with a shared chunk-causal flow backbone.
It organizes text blocks and images as ordered continuous hyperchunks, preserving textual token order and visual spatial structure.
The backbone learns a single vector field over these hyperchunks through Flow Matching.
Joint attention enables cross-modal interaction, while modality-specific feed-forward networks process each modality.
The model predicts multiple target chunks in parallel during training and generates hyperchunks sequentially at inference.
We instantiate \textbf{MF-1} and pretrain it on multimodal data.
Across 0.6B, 1.2B, and 1.6B scales, continued pretraining consistently improves multimodal modeling. With only 150B pretraining tokens, MF-1 achieves an average score of 82.8 across GenEval and DPG-Bench and 75.3 across VQAv2, MMBench, and POPE, remaining competitive with unified models trained on substantially more data. 
Under matched data, optimization, and parameter budgets, Multimodal Flow further outperforms representative hybrid and discrete models.
These results establish continuous chunk-based embedding flow modeling as a new fully continuous paradigm for unified multimodal modeling. 
The related code and model are publicly released at
\href{https://github.com/hustvl/Multimodal-Flow}
{\textcolor{blue}{github.com/hustvl/Multimodal-Flow}}.

\end{abstract}

\section{Introduction}
\label{sec:introduction}


Vision-language models have advanced visual understanding by conditioning text generation on images \citep{liu2023visual,taoinfinitevl,zeng2026diffusionvl}. 
Recent unified multimodal models extend this setting by treating both text and images as generation targets within a single pretrained model \citep{team2024chameleon,zhou2025transfusion,wang2024emu3,zou2025omnimamba}. 
These advances raise a broader question: \emph{can language and vision share a generative process while preserving the representational fidelity and structure each requires?} 
Doing so is nontrivial because language is organized as an ordered sequence of tokens, whereas images have dense spatial structure, and the two have traditionally relied on different representations and generation mechanisms.

\begin{figure}[!t]
  \centering
  \includegraphics[width=\textwidth]{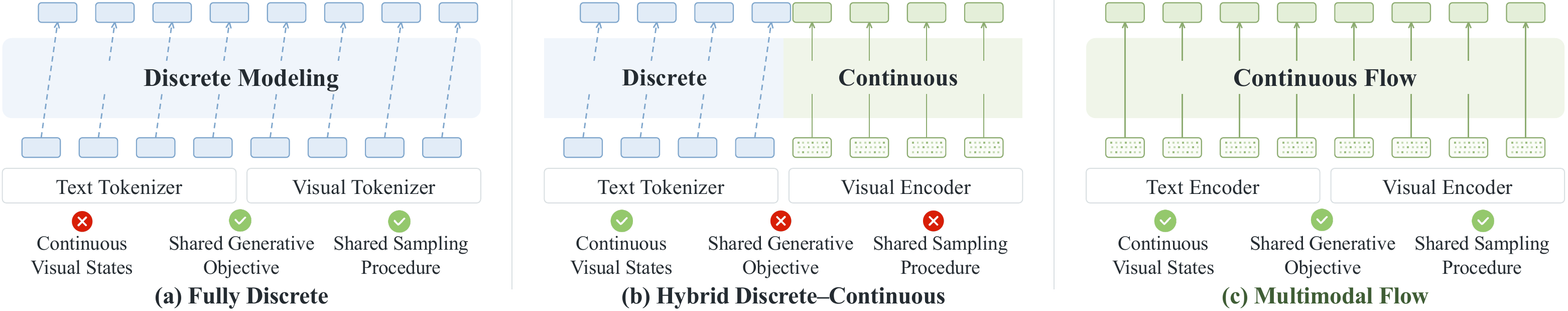}
  \caption{
    \textbf{Comparison of multimodal modeling paradigms.}
    Multimodal Flow combines continuous states with a shared
    generative objective and sampling procedure for language and vision.
    Blue and green denote discrete and continuous states respectively.
    }
  \label{fig:modeling-paradigms}
\end{figure}

Current unified models resolve this tension through two dominant paradigms.
Fully discrete models quantize images into visual tokens, placing both modalities in a shared categorical sequence \citep{team2024chameleon,wang2024emu3,xie2025show}.
This alignment facilitates joint modeling, but makes visual fidelity dependent on the tokenizer: fine-grained details discarded by the tokenizer are unavailable to the model
\citep{tang2026stablevq}.
Hybrid discrete--continuous models instead retain continuous visual states and combine discrete language prediction with diffusion or flow for images \citep{zhou2025transfusion,ma2025janusflow,li2025dual}.
They avoid visual quantization and permit interaction within a shared backbone, but language and vision remain governed by different objectives and sampling procedures.
As summarized in Figure~\ref{fig:modeling-paradigms}, neither paradigm simultaneously provides continuous visual states and a common generative process across modalities.
This motivates fully continuous modeling in embedding spaces, where language and vision can share a continuous objective and sampling mechanism while retaining modality-specific representations.


The technical ingredients for this paradigm are now available.
Continuous diffusion and flow models are well established for visual generation \citep{esser2024scaling,lipman2022flow, ho2020denoising}, and Embedded Language Flows (ELF) shows that contextualized text embeddings can also be modeled directly with Flow Matching \citep{hu2026elf}.
Prior multimodal diffusion and flow systems further demonstrate continuous generation along multiple image--text paths \citep{bao2023one,li2025omniflow, he2025flowtok}.
These systems largely center on joint denoising or predefined cross-modal routes.
A framework that causally factorizes task-defined multimodal sequences and learns from text-only, image-only, and bidirectional paired data through the same continuous pretraining objective remains underexplored.


A direct approach would concatenate continuous text and visual embeddings and apply a shared flow model.
Concatenation alone, however, leaves three questions unanswered: what constitutes a generative unit for each modality, how those units form an ordered conditional process, and how the model can share cross-modal interaction while respecting modality-specific representation statistics.
We address these questions with \textbf{Multimodal Flow}.
First, modality-specific encoders map text blocks and images to continuous embeddings, forming hyperchunks that preserve text-token order and visual-grid structure, respectively.
These heterogeneous hyperchunks provide common units of conditional generation that can be arranged into task-defined multimodal sequences.
Second, a shared chunk-causal flow backbone conditions each target on preceding chunks while modeling positions within the target jointly.
Third, joint attention supports cross-modal interaction, while modality-specific feed-forward networks adapt computation to each representation space.
The corresponding decoders map generated states back to text or images.
Multimodal Flow thus shares generative dynamics and cross-modal interaction without forcing language and vision into a homogeneous representation. Figure~\ref{fig:mixed-multimodal-pretraining} shows parallel target-chunk prediction during training and sequential generation at inference. The same backbone and objective support mixed multimodal pretraining and finetuning for visual question answering and text-to-image generation.

We instantiate the framework as \textbf{MF-1} and evaluate its pretraining, competitive performance, and transfer.
Across 0.6B–1.6B variants, longer pretraining and larger models generally improve language modeling, image captioning, and text-to-image generation.
With 150B pretraining tokens, the 1.6B MF-1 achieves competitive generation and understanding performance against established unified models, scoring 0.821 on GenEval, 83.44 on DPG-Bench, and an average of 75.3 across VQAv2, MMBench and POPE. Controlled comparisons also show that Multimodal Flow outperforms fully discrete and hybrid architectures on the evaluated multimodal tasks. Separately, under matched total training-token budgets of the 1.6B architecture, mixed pretraining raises SeedBench from 31.6 to 62.4 and MMBench from 36.0 to 67.2   relative to a randomly initialized flow backbone, demonstrating MF-1’s ability to transfer learned multimodal representations to downstream tasks.
Additional analyses favor semantic visual representations and modality-specific feed-forward networks, and show that the same classifier-free guidance (CFG) \citep{ho2022classifier} mechanism can improve both text-to-image and image-conditioned text generation.
Together, these results support the feasibility, competitiveness, and transferability of continuous chunk-based embedding modeling on the evaluated language--image tasks.
Additional related work is discussed in Appendix~\ref{sec:related-work}.

In summary, our main contributions are as follows:
\begin{itemize}
    \item We introduce Multimodal Flow, a fully continuous framework that models language and vision in their respective embedding spaces under one Flow Matching objective.
    \item We develop ordered hyperchunks and a chunk-causal backbone that preserve modality-specific structure, support task-defined multimodal sequences, and enable parallel target prediction followed by sequential generation.
    \item We validate MF-1 through scaling, matched architecture comparisons, and downstream transfer, demonstrating the strong multimodal modeling capabilities of Multimodal Flow.
\end{itemize}

\section{Multimodal Flow}
\label{sec:multimodal-flow}

\subsection{Overview}
\label{sec:method-overview}

\begin{figure}[!t]
  \centering
  \includegraphics[width=\textwidth]{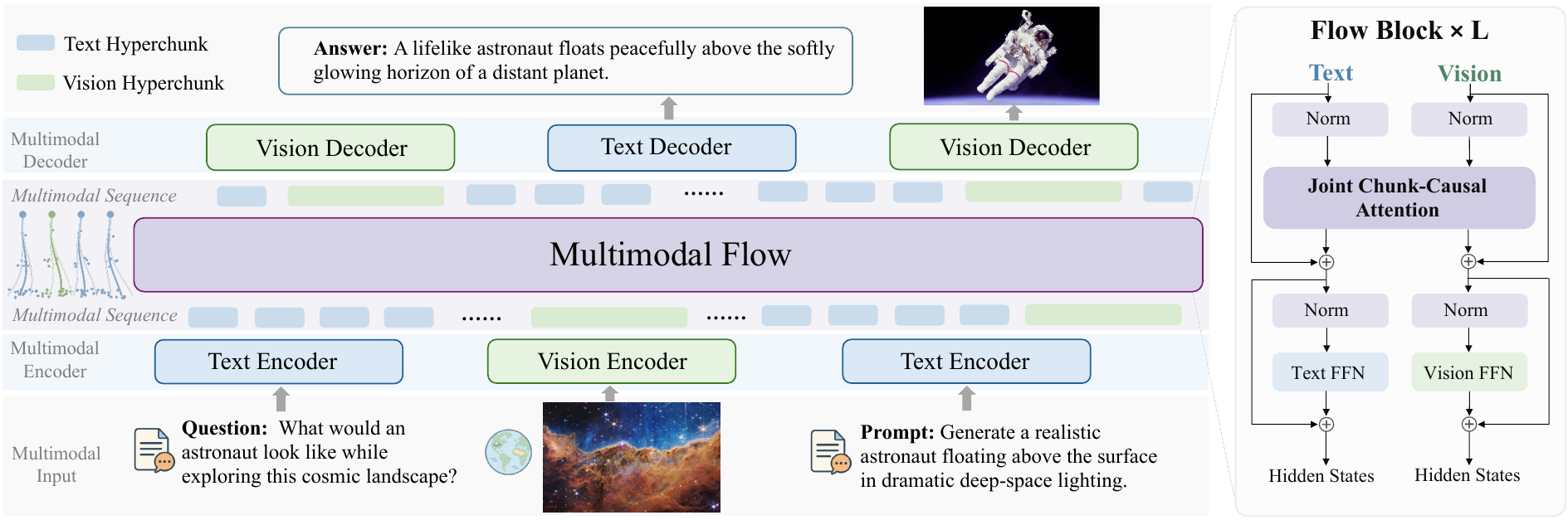}
  \caption{\textbf{Multimodal Flow architecture and ordered hyperchunk representation.} Frozen multimodal encoders map text blocks and images to continuous hyperchunks. After normalization and projection, a shared chunk-causal backbone applies joint attention and modality-specific FFNs. Multimodal decoders map generated hyperchunks back to text or images.}
  \label{fig:multimodal-flow-architecture}
\end{figure}

Figure~\ref{fig:multimodal-flow-architecture} gives an overview of Multimodal Flow. The remainder of this section follows the model dataflow.
We first describe the continuous representations and chunk construction, then introduce the chunk-causal architecture and Flow Matching objective, and finally present parallel training and sequential generation. Section~\ref{sec:training} instantiates the same formulation for mixed multimodal pretraining and downstream finetuning. 

\subsection{Continuous Multimodal Representations}
\label{sec:continuous-multimodal-representations}

Multimodal Flow uses pretrained modality-specific representation encoders to map images and discrete text into continuous states. Given an image $I$ and a text token sequence $Y=(y_1,\ldots,y_{L_\ell})$, we first partition the text into contiguous blocks of length $B$,
\begin{equation}
  Y_b = Y_{(b-1)B+1:\min(bB,L_\ell)},
  \label{eq:text-blocks}
\end{equation}
and encode each block independently. The normalized visual and text representations are
\begin{equation}
  \mathbf{x}^{v}
  =
  \operatorname{Norm}_{v}\!\left(E_v(I)\right),
  \qquad
  \mathbf{x}^{\ell}_{b}
  =
  \operatorname{Norm}_{\ell}\!\left(E_\ell(Y_b)\right).
  \label{eq:multimodal-representations}
\end{equation}
Here, $E_v$ and $E_\ell$ denote the visual and text encoders, while $\operatorname{Norm}_v$ and $\operatorname{Norm}_\ell$ denote the corresponding normalization operations. The two representations retain their respective dimensionalities and internal structures. Modality-specific input projections subsequently map them to a common hidden space.

The continuous states are organized into hyperchunks according to the structure of each modality. Each image forms a visual chunk that preserves its spatial layout, and each independently encoded text block forms a text chunk:
\begin{equation}
  \mathbf{c}^{v}=\mathbf{x}^{v},
  \qquad
  \mathbf{c}^{\ell}_{b}=\mathbf{x}^{\ell}_{b}.
  \label{eq:multimodal-chunks}
\end{equation}
Text chunks retain the original token order. We use $B=8$ in our main models.

After generation, the predicted representations are denormalized and decoded into images or text:
\begin{equation}
  \widehat I
  =
  D_v\!\left(
  \operatorname{Norm}_{v}^{-1}(\widehat{\mathbf{x}}^{v})
  \right),
  \qquad
  \widehat Y_b
  =
  D_\ell\!\left(
  \operatorname{Norm}_{\ell}^{-1}(\widehat{\mathbf{x}}^{\ell}_{b})
  \right).
  \label{eq:modality-decoding}
\end{equation}
In our implementation, we use frozen SigLIP~2 \citep{tschannen2025siglip} and T5-small \citep{raffel2020exploring} encoders, together with a pretrained image decoder \citep{tong2026scaling} and a separately trained text decoder. Both decoders remain frozen during flow pretraining. Representation and decoder details are provided in the appendix.

\subsection{Chunk-Causal Multimodal Flow}
\label{sec:chunk-causal-flow}

\subsubsection{Chunk-Causal Flow Modeling}

Multimodal Flow represents language and vision as an ordered sequence of chunks,
\begin{equation}
  \mathcal C=(\mathbf c_1,\ldots,\mathbf c_K),
  \label{eq:chunk-sequence}
\end{equation}
and factorizes their joint distribution according to the chunk order:
\begin{equation}
  p_\theta(\mathcal C)
  =
  \prod_{k=1}^{K}
  p_\theta\!\left(\mathbf c_k\mid\mathbf c_{<k}\right).
  \label{eq:chunk-factorization}
\end{equation}
When predicting chunk $\mathbf c_k$, all preceding chunks are visible and future chunks remain hidden, while positions within the current chunk are modeled jointly. The chunk order and target modality are specified by the task, allowing the same backbone to operate under different multimodal contexts.

Let $\mathbf x^{(k)}$ denote the clean representation of the target chunk. We construct a linear probability path \citep{lipman2022flow}:
\begin{equation}
  \mathbf z_t^{(k)}
  =
  t\mathbf x^{(k)}
  +
  (1-t)\boldsymbol\epsilon^{(k)},
  \qquad
  \boldsymbol\epsilon^{(k)}\sim\mathcal N(0,\mathbf I),
  \label{eq:chunk-flow-path}
\end{equation}
where $t=0$ corresponds to noise and $t=1$ corresponds to the clean representation. Given the preceding chunks and the perturbed target state, the chunk-causal flow backbone predicts the clean endpoint:
\begin{equation}
  \widehat{\mathbf x}_{\theta}^{(k)}
  =
  f_{\theta}^{m_k}\!\left(
  \mathbf z_t^{(k)},t
  \mid
  \mathbf c_{<k}
  \right),
  \label{eq:chunk-clean-prediction}
\end{equation}
where $m_k$ denotes the modality of the target chunk.

Visual and text states are mapped to a common hidden space through modality-specific input projections and augmented with modality embeddings and timestep embeddings. Multimodal rotary position embedding (MRoPE) \citep{wang2024qwen2} encodes both the chunk order and the positional structure within each chunk. The resulting states are processed by joint self-attention under a chunk-causal mask, allowing information from all preceding visual and text chunks to contribute to the target prediction. The attention outputs are then processed by modality-specific feed-forward networks and projected back to their respective continuous representation spaces.

\subsubsection{Parallel Flow Matching Training}

\begin{figure}[!t]
  \centering
  \includegraphics[width=\textwidth]{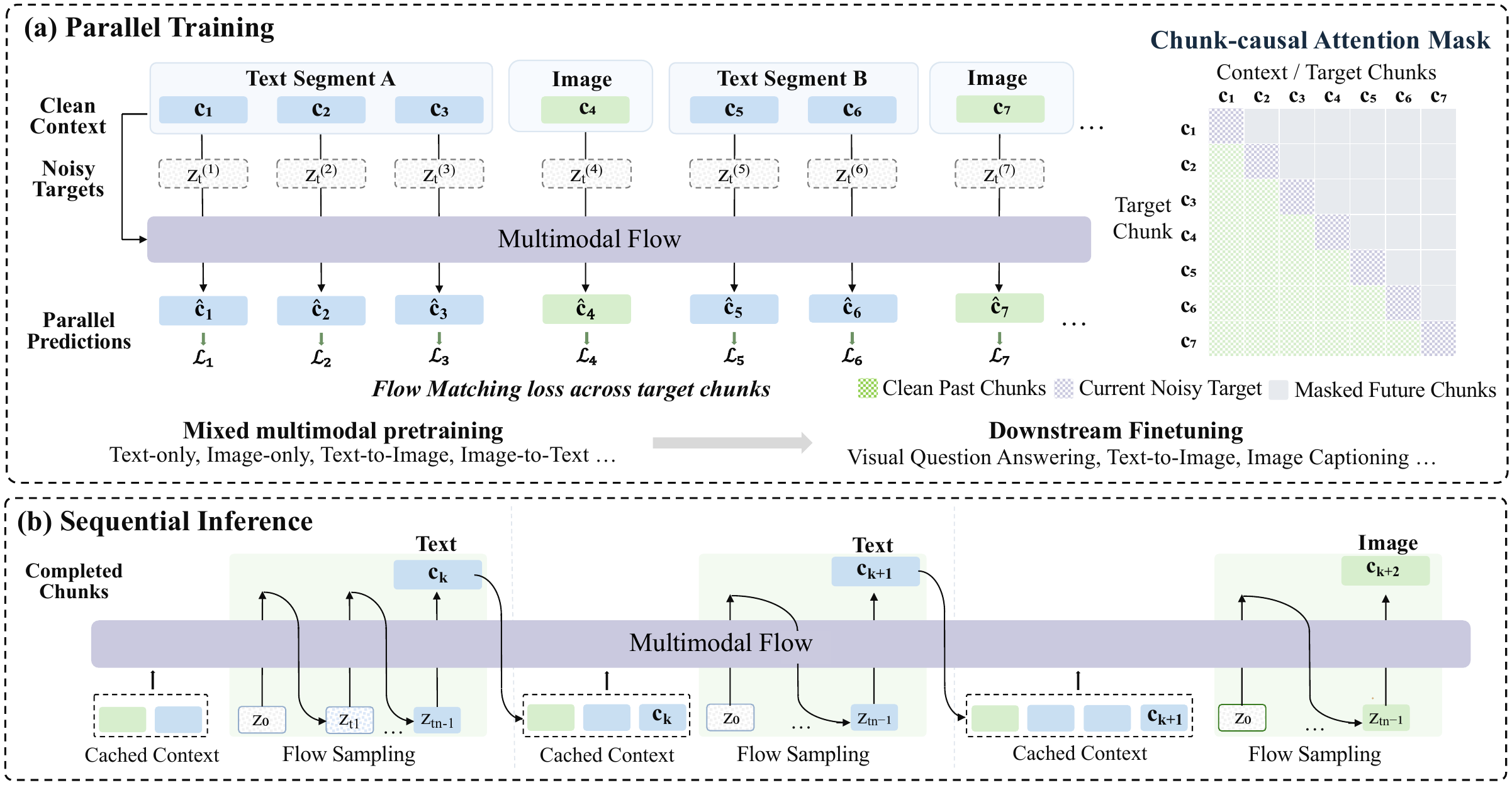}
  \caption{\textbf{Parallel training, mixed multimodal pretraining, and sequential inference.} The chunk-causal formulation predicts multiple target chunks in parallel during training and generates chunks sequentially at inference. Different chunk sequences express unimodal modeling, cross-modal generation, and downstream finetuning within the same Flow Matching objective.}
  \label{fig:mixed-multimodal-pretraining}
\end{figure}

During training, the chunk-causal mask allows multiple target chunks to be predicted in parallel. For a sequence containing multiple text blocks, we construct clean and perturbed views of each block. A perturbed target block can attend to all preceding clean chunks and its own perturbed state, but not to its clean counterpart or any future chunk. Multiple chunk predictions can therefore be computed in a single forward pass. Each target chunk receives an independently sampled flow timestep. Figure~\ref{fig:mixed-multimodal-pretraining}(a) illustrates this parallel construction as part of the overall training procedure.

The model predicts the clean endpoint and converts it into the corresponding velocity along the probability path:
\begin{equation}
  \widehat{\mathbf v}_{\theta}^{(k)}
  =
  \frac{
  \widehat{\mathbf x}_{\theta}^{(k)}-\mathbf z_t^{(k)}
  }{1-t},
  \qquad
  \mathbf v^{(k)}
  =
  \mathbf x^{(k)}-\boldsymbol\epsilon^{(k)}.
  \label{eq:chunk-velocity}
\end{equation}
Let $\mathcal K$ denote the set of target chunks predicted in parallel, and let $\mathbf M^{(k)}$ denote the valid-position mask for chunk $k$. The training objective is
\begin{equation}
  \mathcal L_{\mathrm{flow}}
  =
  \mathbb E\!\left[
  \frac{1}{\sum_{k\in\mathcal K}\lvert\mathbf M^{(k)}\rvert}
  \sum_{k\in\mathcal K}\sum_i
  M_i^{(k)}
  \left\|
  \widehat{\mathbf v}_{\theta,i}^{(k)}-\mathbf v_i^{(k)}
  \right\|_2^2
  \right].
  \label{eq:chunk-flow-loss}
\end{equation}
To improve computational efficiency for variable-length multimodal sequences, we use sequence packing to place multiple logical chunk sequences into a packed sequence with a fixed maximum length. The chunk-causal mask uses sequence identifiers to isolate different samples and prevent information exchange across sequence boundaries. Details of timestep sampling, endpoint handling, and sequence packing are provided in the appendix.

\subsubsection{Sequential Chunk Inference}

As illustrated in Figure~\ref{fig:mixed-multimodal-pretraining}(b), the model generates chunks sequentially in the order specified by the task. Given a clean chunk prefix, the next target chunk is initialized from Gaussian noise. The model then integrates the learned vector field from $t=0$ to $t=1$, conditioned on the preceding chunks. Once generated, the chunk is appended to the context before the model proceeds to the next chunk.

Chunk-causal attention ensures that the hidden states of completed chunks do not depend on future chunks, which enables KV caching. The keys and values of preceding clean chunks are cached at each layer. At each sampling step, only the current target chunk needs to be recomputed. Once generated, its clean state is added to the cache for subsequent chunks.

For conditional generation, we apply CFG. Let $\widehat{\mathbf x}_{\theta,c}$ and $\widehat{\mathbf x}_{\theta,\emptyset}$ denote the predictions obtained with and without the conditioning chunks, respectively. The guided prediction is
\begin{equation}
  \widehat{\mathbf x}_{\theta,\gamma}
  =
  \widehat{\mathbf x}_{\theta,\emptyset}
  +
  \gamma
  \left(
  \widehat{\mathbf x}_{\theta,c}-\widehat{\mathbf x}_{\theta,\emptyset}
  \right),
  \label{eq:chunk-cfg}
\end{equation}
where $\gamma$ is the guidance scale. By specifying different chunk prefixes and target modalities, the same sequential generation process supports both unimodal and cross-modal inference.


\subsection{Mixed Multimodal Pretraining and Downstream Finetuning}
\label{sec:training}

\subsubsection{Mixed Multimodal Pretraining}

The chunk representation converts heterogeneous data sources into task-defined sequences with a common target interface. For text-only data, consecutive text blocks form a sequence $(\mathbf c_1^\ell,\ldots,\mathbf c_K^\ell)$, and every chunk is a target conditioned on its clean prefix; the first chunk is therefore unconditional. An image-only example contains a target visual chunk with no conditioning chunk. For paired image--text data, we use both modality orders: text chunks followed by a visual target for text-to-image generation, and a visual chunk followed by text targets for image-to-text generation. Figure~\ref{fig:mixed-multimodal-pretraining} lists these training configurations.

We pretrain one chunk-causal flow backbone on a mixture of these tasks. Let $\tau$ be a task sampled from mixture distribution $\pi$, let $s\sim\mathcal D_\tau$ be a sample from its data source, and let $\mathcal G_\tau(s)$ construct the corresponding chunk sequence and target set. The mixed-pretraining objective is
\begin{equation}
  \mathcal L_{\mathrm{pre}}
  =
  \mathbb E_{\tau\sim\pi,\,s\sim\mathcal D_\tau}
  \left[
  \mathcal L_{\mathrm{flow}}\!\left(\mathcal G_\tau(s)\right)
  \right].
  \label{eq:mixed-pretraining-objective}
\end{equation}
Across tasks, the continuous interfaces and flow backbone remain fixed under the same Flow Matching objective. Only chunk content, ordering, and target modalities vary. The resulting mixture learns unimodal distributions and bidirectional cross-modal conditionals within a single model.

\subsubsection{Downstream Finetuning}

Downstream finetuning changes the task sequence and data, but not the model interface or objective. For visual question answering, a visual chunk and question chunks form the clean prefix, while answer chunks are text targets. For text-to-image generation, prompt chunks form the prefix and the image is the visual target. The frozen modality codecs are reused, and the chunk-causal backbone is initialized from mixed pretraining and optimized with the same $\mathcal L_{\mathrm{flow}}$.

This separation between task specification and generative modeling is central to Multimodal Flow. Task-specific data and optimization adapt the pretrained backbone, while the chunk interface, causal factorization, and continuous objective remain the same across understanding and generation tasks.

\FloatBarrier

\section{Experiments}
\label{sec:experiments}

\subsection{Experimental Setup}
\label{sec:experiments:setup}

We evaluate MF-1 on language modeling, image captioning, multimodal understanding, and image generation. Our main model has a 1.6B parameter flow backbone trained from scratch with 150B tokens. All MF-1 results in Tables~\ref{tab:geneval_comparison}--\ref{tab:understanding_comparison} use the same checkpoint after 5B tokens of joint finetuning. We also compare 0.6B, 1.2B, and 1.6B variants under a common pretraining and evaluation protocol. Appendices~\ref{app:implementation-details} and~\ref{app:experimental-details} provide the model and codec configurations, training budgets, benchmarks, and evaluation sampling settings.

\subsection{Comparison across Multimodal Modeling Paradigms}
\label{sec:experiments:comparison}

We evaluate MF-1 through two complementary comparisons. The first uses published results from generation-specific, understanding-specific, and unified models under their reported training settings. The second compares modeling paradigms with matched data and optimization schedules under the same trainable-parameter budget.

\begin{table}[!t]
\caption{
Category-level comparison on GenEval.
Unified models are grouped by multimodal modeling paradigm.
Best and second-best results are shown in bold and underlined respectively.
}
\vspace{2pt}
\label{tab:geneval_comparison}
\centering
\footnotesize
\setlength{\tabcolsep}{2pt}
\renewcommand{\arraystretch}{1.06}

\begin{tabularx}{\textwidth}{
  @{}l>{\raggedright\arraybackslash}X*{8}{c}@{}
}
\toprule
Type & Model & Params & Single & Two & Count. &
Colors & Pos. & \mbox{Color Attr.} &
\mbox{Overall$\uparrow$} \\
\midrule

\multirow{3}{*}{\shortstack{\textit{Gen.}\\\textit{only}}}
& SDXL~\citep{podell2024sdxl}
& 2.6B
& \underline{0.98}
& 0.74 & 0.39 & 0.85 & 0.15 & 0.23 & 0.55 \\

& Hunyuan-DiT~\citep{li2024hunyuan}
& 1.5B
& 0.97
& 0.77
& \underline{0.71}
& \underline{0.88}
& 0.13 & 0.30 & 0.63 \\

& SD3 Medium~\citep{esser2024scaling}
& 2.0B
& \underline{0.98}
& 0.74 & 0.63 & 0.67 & 0.34 & 0.36 & 0.62 \\

\midrule

\multirow{12}{*}{\textit{Unified}}
& \multicolumn{9}{l}{\textit{Fully discrete}} \\

& LWM~\citep{liu2025world}
& 7B
& 0.93 & 0.41 & 0.46 & 0.79 & 0.09 & 0.15 & 0.47 \\

& Show-o~\citep{xie2025show}
& 1.3B
& 0.95 & 0.52 & 0.49 & 0.82 & 0.11 & 0.28 & 0.53 \\

& Emu3-Gen~\citep{wang2024emu3}
& 8B
& \underline{0.98}
& 0.71 & 0.34 & 0.81 & 0.17 & 0.21 & 0.54 \\

& Muddit~\citep{shi2025muddit}
& 1B
& \underline{0.98}
& 0.72 & 0.54 & 0.82 & 0.19 & 0.41 & 0.61 \\

& UniDisc~\citep{swerdlow2025unified}
& 1.4B
& 0.92 & 0.47 & 0.15 & 0.67 & 0.13 & 0.19 & 0.42 \\

\cmidrule(lr){2-10}
& \multicolumn{9}{l}{\textit{Hybrid discrete--continuous}} \\

& Transfusion~\citep{zhou2025transfusion}
& 7B
& -- & -- & -- & -- & -- & -- & \underline{0.67} \\

& JanusFlow~\citep{ma2025janusflow}
& 1.3B
& 0.97 & 0.59 & 0.45 & 0.83
& \underline{0.53}
& 0.42 & 0.63 \\

& D-DiT~\citep{li2025dual}
& 2B
& 0.97
& \underline{0.80}
& 0.54 & 0.76 & 0.32
& \underline{0.50}
& 0.65 \\

\cmidrule(lr){2-10}
& \multicolumn{9}{l}{\textit{Fully continuous}} \\

\rowcolor{gray!10}
& \textbf{MF-1}
& 1.6B
& \textbf{0.99}
& \textbf{0.92}
& \textbf{0.73}
& \textbf{0.90}
& \textbf{0.81}
& \textbf{0.58}
& \textbf{0.82} \\

\bottomrule
\end{tabularx}
\end{table}

\begin{table}[!t]
\caption{Detailed text-to-image generation performance on DPG-Bench.}
\label{tab:dpg_comparison}
\centering
\small
\setlength{\tabcolsep}{3pt}
\renewcommand{\arraystretch}{1.06}

\begin{tabularx}{\textwidth}{
  @{}lc
  *{5}{>{\hsize=.94\hsize
         \linewidth=\hsize
         \centering\arraybackslash}X}
  >{\hsize=1.30\hsize
    \linewidth=\hsize
    \centering\arraybackslash}X
  @{}
}
\toprule
Model & Params & Global & Entity & Attribute &
Relation & Other & \mbox{Overall$\uparrow$} \\
\midrule

SDXL~\citep{podell2024sdxl}
& 2.6B
& 83.27 & 82.43 & 80.91 & 86.76 & 80.41 & 74.65 \\

Hunyuan-DiT~\citep{li2024hunyuan}
& 1.5B
& 84.59 & 80.59 & 88.01 & 74.36 & 86.41 & 78.87 \\

SD3 Medium~\citep{esser2024scaling}
& 2.0B
& \underline{87.90}
& \textbf{91.01}
& \underline{88.83}
& 80.70
& \underline{88.68}
& \textbf{84.08} \\

Show-o~\citep{xie2025show}
& 1.3B
& 79.33 & 75.44 & 78.02 & 84.45 & 60.80 & 67.27 \\

Janus~\citep{wu2025janus}
& 1.3B
& 82.33 & 87.38 & 87.70 & 85.46 & 86.41 & 79.68 \\

Emu3-Gen~\citep{wang2024emu3}
& 8B
& 85.21 & 86.68 & 86.84
& \textbf{90.22}
& 83.15 & 80.60 \\

JanusFlow~\citep{ma2025janusflow}
& 1.3B
& 87.03 & 87.31 & 87.39
& \underline{89.79}
& 88.10 & 80.09 \\

\rowcolor{gray!10}
\textbf{MF-1}
& 1.6B
& \textbf{89.38}
& \underline{89.03}
& \textbf{90.82}
& 89.73
& \textbf{89.47}
& \underline{83.44} \\

\bottomrule
\end{tabularx}
\end{table}

\begin{table}[!t]
\caption{
Multimodal understanding performance and pretraining scale.
PT Tok. denotes cumulative pretraining tokens along the backbone
inheritance chain.
}

\label{tab:understanding_comparison}
\centering
\footnotesize
\setlength{\tabcolsep}{1.6pt}
\renewcommand{\arraystretch}{1.06}

\newcommand{\tablecite}[1]{{\scriptsize\citep{#1}}}
\begin{tabularx}{\textwidth}{
  @{}l
  >{\raggedright\arraybackslash}X
  *{8}{c}
  @{}
}
\toprule
Paradigm & Model & Params & \mbox{PT Tok.} &
POPE & MMB & SEEDB & VQAv2 & GQA & \mbox{OK-VQA} \\
\midrule

\multicolumn{10}{l}{
\textit{With Pretrained LLM Initialization}
} \\
\addlinespace[2pt]

\multirow{4}{*}{\textit{Und. only}}
& \mbox{MobileVLM-V2~\tablecite{chu2024mobilevlm}}
& 2.7B & $\sim$1.3T
& 84.7 & 63.2 & -- & -- & 61.1 & -- \\

& LLaVA-Phi~\citep{zhu2024llava}
& 2.7B & $\sim$1.4T
& 85.0 & 59.8 & -- & 71.4 & -- & -- \\

& LLaVA-v1.5~\citep{liu2024improved}
& 7.0B & $\sim$2.0T
& 85.9 & 64.3 & 58.6 & 78.5 & 62.0 & -- \\

& \mbox{Qwen-VL-Chat~\tablecite{bai2023qwen}}
& 7.0B & $\sim$3.1T
& 83.7 & 60.6 & 58.2 & 78.2 & 57.5 & 56.6 \\

\addlinespace[2pt]

\multirow{3}{*}{\textit{Discrete}}
& Janus~\citep{wu2025janus}
& 1.3B & $\sim$0.9T
& 87.0 & 69.4 & 63.7 & 77.3 & 59.1 & -- \\

& Show-o~\citep{xie2025show}
& 1.3B & $\sim$1.0T
& 73.8 & -- & -- & 59.3 & 48.7 & -- \\

& Janus-Pro~\citep{chen2025janus}
& 1.5B & $\sim$1.1T
& 86.2 & 75.5 & 68.3 & -- & 59.3 & -- \\

\addlinespace[2pt]

\multirow{2}{*}{\textit{Hybrid}}
& JanusFlow~\citep{ma2025janusflow}
& 1.3B & $\sim$2.0T
& 88.0 & 74.9 & 70.5 & 79.8 & 60.3 & -- \\

& Show-o2~\citep{xie2026show}
& 1.5B & $\sim$18.2T
& 83.0 & 67.4 & 65.6 & -- & 60.0 & -- \\

\midrule

\multicolumn{10}{l}{
\textit{Without Pretrained LLM Initialization}
} \\
\addlinespace[2pt]

\textit{Discrete}
& Muddit~\citep{shi2025muddit}
& 1.0B & $\sim$400B
& 52.5 & 28.4 & 13.0 & 68.2 & 57.5 & 21.1 \\

\textit{Hybrid}
& D-DiT~\citep{li2025dual}
& 2.0B & $\sim$4.3T
& 79.2 & -- & -- & 59.5 & 55.1 & 28.5 \\

\rowcolor{gray!10}
\textit{Continuous}
& \textbf{MF-1}
& 1.6B & 150B
& 86.1 & 67.2 & 62.4 & 72.6 & 58.3 & 39.1 \\

\bottomrule
\end{tabularx}
\end{table}

The image-generation comparisons show that MF-1 achieves the strongest compositional generation performance and the best long-prompt alignment among unified models. This indicates that jointly modeling language and images within a continuous generative process produces language representations that transfer effectively to text-conditioned image synthesis. MF-1 also performs strongly on multimodal understanding despite being trained from scratch on only 150B tokens. It consistently surpasses comparable from-scratch unified models such as Muddit and D-DiT, while remaining competitive with similarly sized models initialized from pretrained LLMs. 

\begin{table*}[!t]
  \caption{Controlled comparison of multimodal architectures under matched data, optimization schedules, and trainable parameter budgets.}
  \label{tab:architecture_comparison}
  \centering
  \small
  \setlength{\tabcolsep}{5pt}
  \renewcommand{\arraystretch}{1.08}

  \resizebox{\textwidth}{!}{%
    \begin{tabular}{@{}llccccc@{}}
      \toprule
      Architecture
        & Architecture Type
        & GenEval $\uparrow$
        & GQA $\uparrow$
        & VQAv2 $\uparrow$
        & MMBench $\uparrow$
        & SEEDB $\uparrow$ \\
      \midrule
      Multimodal Flow
        & Fully Continuous
        & \textbf{0.7134}
        & \textbf{55.60}
        & \textbf{69.03}
        & \textbf{46.74}
        & \textbf{51.85} \\
      Transfusion-Style
        & Discrete--Continuous Hybrid
        & 0.6693
        & 52.81
        & 68.49
        & 33.68
        & 31.31 \\
      Chameleon-Style
        & Fully discrete
        & 0.3744 & 45.83 & 56.37 & 38.40 & 42.40 \\
      \bottomrule
    \end{tabular}%
  }
\end{table*}

To assess Multimodal Flow under a matched training budget, we compare it with fully discrete and hybrid architectures using the same data, optimization schedule, and trainable-parameter budget. The Transfusion-style hybrid provides a particularly close reference, sharing MF-1’s visual pathway and backbone design while using autoregressive text modeling. MF-1 achieves stronger image generation and visual understanding across these comparisons (Table~\ref{tab:architecture_comparison}), establishing it as an effective fully continuous architecture for unified multimodal modeling.

\subsection{Mixed Multimodal Pretraining and Model Capacity}
\label{sec:experiments:mixed-pretraining}

Figure~\ref{fig:multimodal-scaling} tracks the 0.6B, 1.2B, and 1.6B variants under the same mixed-pretraining and evaluation protocol, reporting the Flow Matching objective, language PPL, GenEval, and image-captioning CLIPScore. All three variants show consistent overall improvement: the Flow Matching objective and language PPL decrease, while GenEval and CLIPScore increase as training proceeds. Increasing model capacity yields clearer gains in language modeling and image captioning, with the 1.6B model maintaining the strongest performance later in training. These trends show that a shared chunk-causal Flow Matching objective can jointly develop language modeling, image-conditioned text generation, and text-to-image generation capabilities, with further gains from longer training and increased model capacity.

\begin{table}[!t]
  \caption{Downstream performance with random and mixed-pretrained initialization.}
  \label{tab:pretrain_transfer}
  \centering
  \small
  \renewcommand{\arraystretch}{1.08}
  \begin{tabular*}{\columnwidth}{@{\extracolsep{\fill}}lccccc@{}}
    \toprule
    Initialization & GenEval $\uparrow$ & DPG-Bench $\uparrow$ & SEEDB $\uparrow$ & MMB $\uparrow$ & OK-VQA $\uparrow$ \\
    \midrule
    Random & 0.527 & 57.81 & 31.6 & 36.0 & 22.9 \\
    Mixed pretraining & \textbf{0.821} & \textbf{83.44} & \textbf{62.4} & \textbf{67.2} & \textbf{39.1} \\
    \bottomrule
  \end{tabular*}
\end{table}

\begin{figure}[!t]
  \vspace{6pt}
  \centering
  \includesvggraphic[width=1.00\textwidth]{figures/wave5_scaling_2x2.svg}
  \vspace{-20pt}
  \caption{Training progress under mixed multimodal pretraining at different model capacities. We compare the Flow Matching objective, GPT-2-large PPL, GenEval, and CLIPScore for the 0.6B, 1.2B, and 1.6B models.}
  \label{fig:multimodal-scaling}
  \vspace{10pt}
  \begin{minipage}[t]{0.49\linewidth}
    \centering
    \includesvggraphic[width=\linewidth]{figures/visual_representation_ablation.svg}
  \end{minipage}\hfill
  \begin{minipage}[t]{0.49\linewidth}
    \centering
    \includesvggraphic[width=\linewidth]{figures/cfg_sweep.svg}
  \end{minipage}
  \vspace{-8pt}
  \caption{Design analysis. (a) Image-generation and image-captioning performance under different visual representation spaces. (b) Performance as a function of classifier-free guidance scale. Outlined marks and value labels indicate the best result for each metric.}
  \label{fig:design-analysis}
\end{figure}

\subsection{Downstream Finetuning from Mixed Pretraining}
\label{sec:experiments:finetuning}

We compare mixed-pretrained and randomly initialized 1.6B models with the same architecture, matching the baseline's downstream training tokens to the pretrained model's combined pretraining and finetuning budget. Table~\ref{tab:pretrain_transfer} shows consistent gains in GenEval, DPG-Bench and all VQA benchmarks, supporting the transfer benefits of mixed pretraining rather than increased token exposure.

\subsection{Design Analysis}
\label{sec:experiments:analysis}

We analyze three design choices in MF-1: the continuous visual representation space, the balance between shared and modality-specific computation, and CFG for text and image generation.

\subsubsection{Continuous Visual Representation Space}
\label{sec:experiments:visual-representation}

The visual representation encoder defines the continuous state modeled by the flow. Figure~\ref{fig:design-analysis}(a) compares DINOv2~\citep{oquab2023dinov2} and SigLIP2~\citep{tschannen2025siglip} representations with FLUX.2 VAE latents~\citep{blackforestlabs2025flux2}, SD-VAE latents~\citep{rombach2022high}, and raw image patches.

At the evaluated training budget, DINOv2 achieves the highest image-generation scores, whereas SigLIP2 provides a better balance between generation and captioning, yielding higher CIDEr and CLIPScore despite lower GenEval and DPG-Bench scores. Reconstruction-oriented VAE representations underperform semantic embeddings on both generation and captioning metrics in this setting. Raw pixel representations also yield weak performance on both tasks.

\subsubsection{Shared Interaction and Modality-Specific Computation}
\label{sec:experiments:parameterization}

\begin{table}[!t]
  \caption{Comparison of cross-modal parameterizations after 50B pretraining tokens.}
  \label{tab:modality_parameterization}
  \centering
  \small
  \renewcommand{\arraystretch}{1.08}
  \begin{tabular*}{\textwidth}{@{\extracolsep{\fill}}llccccc@{}}
    \toprule
    Attn. projection & FFN & Text PPL $\downarrow$ & GenEval $\uparrow$ & DPG $\uparrow$ & CIDEr $\uparrow$ & CLIPScore $\uparrow$ \\
    \midrule
    Specific & Specific & 28.09 & \textbf{0.255} & 71.01 & \textbf{48.14} & \textbf{0.820} \\
    Specific & Shared & 29.67 & 0.233 & 70.82 & 47.31 & \textbf{0.820} \\
    Shared & Specific & \textbf{27.43} & 0.226 & \textbf{71.04} & 47.91 & \textbf{0.820} \\
    Shared & Shared & 28.35 & 0.237 & 69.70 & 42.02 & 0.788 \\
    \bottomrule
  \end{tabular*}
\end{table}



MF-1 uses joint attention for cross-modal interaction, with either shared or modality-specific attention projections and FFNs. Table~\ref{tab:modality_parameterization} compares four parameter-matched configurations. Shared and modality-specific attention projections perform similarly overall, whereas sharing FFNs degrades performance, particularly on image-conditioned text generation. These results motivate shared attention projections for cross-modal interaction and modality-specific FFNs to accommodate the distinct representation statistics of language and vision.

\subsubsection{Bidirectional Classifier-Free Guidance}
\label{sec:experiments:cfg}

MF-1 applies the same CFG mechanism to text-to-image and image-conditioned text generation. We fix the pretrained EMA checkpoint of the 0.6B model and vary only the guidance scale during inference.

As shown in Figure~\ref{fig:design-analysis}(b), image captioning performs best at a guidance scale of 3, while text-to-image generation reaches its strongest performance at a scale of 5 and begins to decline as the scale increases further. These results show that the same continuous flow formulation supports CFG for both text and image generation, with the guidance strength adjusted according to the output modality.

\FloatBarrier

\section{Conclusion}
\label{sec:conclusion}

We introduced \textbf{Multimodal Flow}, a fully continuous model for language--vision pretraining, instantiated as \textbf{MF-1}. 
It organizes text blocks and images as ordered hyperchunks and models them with a shared chunk-causal flow backbone, preserving modality-specific structure within a common generative process. 
The resulting framework expresses diverse language--vision tasks as ordered hyperchunk sequences, unifying multimodal understanding and generation under a single continuous objective. 
Experiments show gains with increased model scale and training tokens, competitive generation and understanding performance, and substantial downstream benefits from mixed pretraining. 
Together, these results establish continuous hyperchunk modeling as a flexible paradigm for multimodal pretraining and show its ability to learn transferable representations across language and vision. 
Extending it to longer interleaved sequences, video, and other structured modalities remains a promising direction.
We leave it as future work.

\section*{Acknowledgments}
We thank Lunbin Zeng and Shuai Zhang for valuable discussions and insightful feedback that contributed to this work.

\bibliography{iclr2027_conference}
\bibliographystyle{iclr2027_conference}

\clearpage
\appendix
\section{Related Work}
\label{sec:related-work}

\subsection{Discrete and Hybrid Discrete--Continuous Multimodal Modeling}
\label{sec:related-discrete-hybrid}

Generative foundation models for language and vision have largely followed two architectural paradigms: fully discrete and hybrid discrete--continuous modeling. Fully discrete models use a visual tokenizer to map images into discrete visual tokens. Text and images can then form a common discrete sequence and be modeled through autoregressive prediction, masked diffusion, or discrete flow \citep{team2024chameleon,wang2024emu3,xie2025show,yang2026mmada,wang2026fudoki}. This design aligns the state type, prediction target, and generative process across modalities, while allowing the model to reuse established language model training recipes. For images, however, representation quality is bounded by the visual tokenizer. Improving reconstruction fidelity generally requires a larger codebook or longer token sequences, while information lost through quantization cannot be fully recovered by the downstream backbone \citep{vtp}.

Hybrid discrete--continuous models preserve discrete language modeling while generating continuous visual representations with diffusion or flow. A representative design retains autoregressive next-token prediction for text and pairs it with a continuous generation objective for images \citep{zhou2025transfusion, deng2025emerging, ma2025janusflow, guo2025visual}. Other systems replace the autoregressive language objective with discrete diffusion or related flow-based objectives \citep{li2025dual, nguyen2025oneflow}. These models avoid visual quantization, but a single system must still coordinate different training objectives, perturbation schedules, and sampling procedures. 

\subsection{Continuous Generative Modeling of Language and Vision}
\label{sec:related-continuous}

Continuous diffusion and flow models are now widely used for visual generation. They typically learn probability paths from noise to data in reconstruction-oriented VAE latent spaces \citep{rombach2022high,peebles2023scalable,lipman2022flow,yao2024fasterdit,zhu2025dig,zhang2025mobilei2v,zhu2026stream}. Representation Autoencoders further incorporate pretrained visual encoders, producing continuous generative states with richer semantic information and spatial structure \citep{zheng2026diffusion,tong2026scaling}. More recently, continuous generative modeling has been extended to language. One line of work operates directly in token-level embedding spaces, while another compresses sentences or text blocks into continuous latents and recovers discrete text with a language decoder \citep{li2022diffusion,gong2022diffuseq,lovelace2023latent,hu2026elf,guo2026continuous}. Together, these studies demonstrate continuous language generation across different representation granularities, ranging from token-level embeddings to compressed sentence- and block-level latents.


Prior studies have established continuous multimodal generation through composable diffusion, joint denoising, and cross-modal flows \citep{xu2023versatile,tang2023any,bao2023one,li2025omniflow,liu2025flowing,he2025flowtok}.
Many nevertheless generate from reconstruction-oriented VAE latents or compact visual tokens rather than semantic representations.
These approaches also tend to couple modalities through joint denoising or direct mappings, without explicitly treating token-ordered text and spatially structured images as units of an ordered conditional sequence.
Meanwhile, mixed pretraining of a randomly initialized continuous backbone on multimodal data remains largely unexplored.
Multimodal Flow addresses these gaps by organizing contextual text embeddings and spatial semantic visual embeddings into ordered hyperchunks and modeling them with one chunk-causal Flow Matching objective.
MF-1's scaling, transfer, and controlled comparisons show that this formulation can serve as a practical foundation for multimodal pretraining beyond cross-modal generation.

\clearpage
\section{Implementation Details}
\label{app:implementation-details}

\subsection{Backbone Configuration}
\label{app:model-configurations}

Table~\ref{tab:appendix-model} summarizes the main 1.6B MF-1 configuration. 

\begin{table}[H]
  \caption{Backbone and representation settings of the main MF-1 model.}
  \label{tab:appendix-model}
  \centering
  \small
  \setlength{\tabcolsep}{6pt}
  \renewcommand{\arraystretch}{1.12}
  \begin{tabularx}{\linewidth}{@{}lXlX@{}}
    \toprule
    Backbone & Setting & Representation & Setting \\
    \midrule
    Parameters & 1.6B & Text encoder & T5-small \\
    Transformer layers & 32 & Text embedding dim. & 512 \\
    Hidden dimension & 1600 & Tokens per text chunk & 8 \\
    Attention heads & 25 & Text input bottleneck & 128 \\
    Head dimension & 64 & Vision encoder & SigLIP2-so400m \\
    FFN hidden dimension & 4224 & Image input resolution & 224 \\
    Attention projections & Shared & Visual grid & $16\times16$ \\
    FFN parameters & Modality-Specific & Visual embedding dim. & 1152 \\
    \bottomrule
  \end{tabularx}
\end{table}

\subsection{Representation Encoders and Decoders}
\label{app:codecs}

\paragraph{Text.}
We pretrain the text decoder separately from the multimodal flow backbone. A frozen T5-small encoder maps text to 512-dimensional latents, which are normalized and detached before decoder training. For each token position, we independently sample $z\sim\mathcal{N}(0,1)$, set $\lambda=\sigma(0.8+0.8z)$, and perturb the clean latent $h$ as
$\widetilde h=\lambda h+(1-\lambda)s\epsilon$,
where $\epsilon\sim\mathcal{N}(0,I)$ and $s$ controls the noise strength. A six-layer bidirectional Transformer of width 512 predicts the original tokens in parallel within each eight-token block. We minimize cross-entropy averaged over target text tokens, excluding conditioning and padding positions. This stage trains only the decoder. The resulting decoder is frozen during multimodal pretraining. At inference, it converts generated text latents into tokens block by block.

\paragraph{Vision.}
The main configuration uses a frozen SigLIP2-so400m encoder~\citep{tschannen2025siglip} with a patch size of 14. Each $224\times224$ image produces 256 patch embeddings of dimension 1152, retained as a single spatially structured chunk. A pretrained representation decoder~\citep{tong2026scaling} reconstructs images from the generated visual embeddings. Both text and visual codecs remain frozen during flow pretraining and downstream finetuning.

\paragraph{Normalization.}
We normalize both modalities before flow modeling. For each active text token, the frozen T5-small encoder produces $h^{\mathrm{text}}_{i,d}$, which is normalized as
$\tilde h^{\mathrm{text}}_{i,d}
=(h^{\mathrm{text}}_{i,d}-\mu_{\mathrm{text}})/\sigma_{\mathrm{text}}$
using dimension-shared constants $\mu_{\mathrm{text}}=0$ and
$\sigma_{\mathrm{text}}=0.2$. EOS tokens are treated identically; padding
positions are set to zero and excluded from the objective. For vision, we
normalize each token position and channel:
$\tilde h^{\mathrm{vis}}_{i,d}
=(h^{\mathrm{vis}}_{i,d}-\mu^{\mathrm{vis}}_{i,d})/
\sigma^{\mathrm{vis}}_{i,d}$.
The $256\times1152$ population moments were estimated from 50,176 GPIC
training images.
Generated vision states are denormalized before image decoding,
$h^{\mathrm{vis}}_{i,d}
=\tilde h^{\mathrm{vis}}_{i,d}\sigma^{\mathrm{vis}}_{i,d}
+\mu^{\mathrm{vis}}_{i,d}$.


\subsection{Flow Training and Sequence Packing}
\label{app:flow-training}

\paragraph{Timestep sampling and endpoint conversion.}
We sample a noise level from a shifted logit-normal distribution:
$z\sim\mathcal N(0,1)$, $u=\operatorname{sigmoid}(z)$,
$q=\alpha u/[1+(\alpha-1)u]$, and $t=1-q$.
We set $\alpha=8$ for image targets and $\alpha=6$ for text targets.
Timesteps are sampled independently for target chunks: one per
256-token image chunk and one per 8-token text block. With $x$ denoting
the clean normalized state, the flow path is $z_t=tx+(1-t)\epsilon$.
For numerical stability near the clean endpoint, the implementation
converts the predicted clean state to velocity using
$\hat v_\theta=(\hat x_\theta-z_t)/\max(1-t,0.05)$.

\paragraph{Sequence packing.}
We pack independent examples into physical sequences of at most
32,768 model positions. Each example retains a distinct sequence
identifier: chunk-causal attention permits the prescribed interactions
within an example but blocks attention across examples in the same pack.
Padding positions are inactive in attention. 

\section{Experimental Protocols}
\label{app:experimental-details}

\subsection{Training Configurations}
\label{app:training-configurations}

We distinguish the main benchmark model from the controlled architecture and design studies. Table~\ref{tab:appendix-budgets} records their model sizes and pretraining and finetuning token budgets. The scaling study follows the 0.6B, 1.2B, and 1.6B variants at matched checkpoints, as shown in Figure~\ref{fig:multimodal-scaling}.

\begin{table}[H]
  \caption{Training token budgets for the main and controlled experiments.
  PT and FT denote pretraining and finetuning respectively.}
  \label{tab:appendix-budgets}
  \centering
  \small
  \setlength{\tabcolsep}{5pt}
  \renewcommand{\arraystretch}{1.12}
  \begin{tabularx}{\linewidth}{@{}Xlccc@{}}
    \toprule
    Experiment & Reference & Model size & PT Tokens & FT Tokens \\
    \midrule
    Main benchmark comparisons
      & Tables~\ref{tab:geneval_comparison}--\ref{tab:understanding_comparison}
      & 1.6B & 150B & 5B \\
    Controlled architectures
      & Table~\ref{tab:architecture_comparison}
      & 1.6B & 50B & 5B \\
    Pretraining transfer
      & Table~\ref{tab:pretrain_transfer}
      & 1.6B & 150B & 5B \\
    Attention/FFN parameterization
      & Table~\ref{tab:modality_parameterization}
      & 0.6B & 50B & 0B \\
    Visual representations and CFG
      & Figure~\ref{fig:design-analysis}
      & 0.6B & 50B & 0B \\
    \bottomrule
  \end{tabularx}
\end{table}

\paragraph{Pretraining data.}
We use GPIC~\citep{chandrasegaran2026gpic} for image generation,
LLaVA-OneVision-1.5~\citep{an2025llava} for image understanding,
and Ultra-FineWeb-L3~\citep{wang2025ultrafineweb} for text-only modeling.
The image-only task uses GPIC images without their paired text.

\paragraph{Finetuning data.}
For text-to-image finetuning, we use BLIP3o-60k~\citep{chen2025blip3},
the LAION DALL-E 3 Discord dataset~\citep{opendatasetsdalle3},
and ShareGPT-4o-Image~\citep{chen2025sharegpt}.
For VQA finetuning, we primarily use the LLaVA-v1.5 visual instruction-tuning data~\citep{liu2024improved}.

\paragraph{Mixed pretraining.}
The task mixture consists of 70\% text-only modeling, 20\% image understanding, 9\% text-to-image generation, and 1\% image-only modeling. The main 1.6B model is pretrained from scratch on 150B tokens, followed by 5B finetuning tokens for downstream evaluation. The same Flow Matching objective is used across unimodal and cross-modal tasks.

\paragraph{Controlled comparisons.}
The architecture comparison uses 1.6B models with 50B pretraining
and 5B finetuning tokens under matched data and optimization settings.
The pretraining-transfer comparison instead uses a 1.6B model with
150B pretraining and 5B finetuning tokens. Its randomly initialized
counterpart is trained directly on downstream data for 155B tokens,
matching the total token budget. The attention/FFN and visual
representation/CFG studies use 0.6B models with 50B pretraining without downstream finetuning.

\paragraph{Architecture-specific configurations.}
Table~\ref{tab:appendix-architecture} summarizes the representations, training objectives, and parameter-sharing choices of the architectures compared in Table~\ref{tab:architecture_comparison}.

\begin{table}[!t]
\caption{Representations and objectives in the controlled architecture comparison. AR denotes autoregressive cross-entropy; flow denotes velocity MSE.
Parenthesized values in the last column are FFN hidden dimensions.}
\label{tab:appendix-architecture}
\centering
\footnotesize
\setlength{\tabcolsep}{3pt}
\renewcommand{\arraystretch}{1.13}
\begin{tabularx}{\linewidth}{@{}l
  >{\raggedright\arraybackslash}X
  >{\raggedright\arraybackslash}X
  >{\raggedright\arraybackslash}X@{}}
\toprule
Architecture & Text & Vision & Attention / FFNs \\
\midrule
Multimodal Flow
& T5 latent (512; 8-token blocks), flow
& SigLIP2 latent ($256\times1152$), flow
& Shared / modality-specific (4224) \\

Chameleon-Style
& T5 token IDs, AR
& Janus VQ-16 (576 IDs; 16,384 codes), AR
& Shared / modality-specific (4000) \\

Transfusion-style
& T5 token IDs, AR
& SigLIP2 latent ($256\times1152$), flow
& Shared / modality-specific (4062) \\

\bottomrule
\end{tabularx}
\end{table}

All trainable backbones are randomly initialized and share 32 layers,
width 1600, 25 attention heads of dimension 64, RMSNorm, and positional
encoding conventions. Their FFN widths are adjusted so that trainable
parameter counts differ from MF-1's 1,634,591,168 by at most 25,536
(0.00156\%); frozen codecs are excluded. The comparison matches the
data stream, supervised-token exposure, optimization recipe, and
trainable-parameter budget, while preserving each architecture's
native objective.

\subsection{Benchmarks and Metrics}
\label{app:benchmarks}

\paragraph{Language modeling and captioning.}
We evaluate generated text using GPT-2-large perplexity~\citep{radford2019language}, an external fluency metric rather than the flow model's likelihood. Image captioning is evaluated on the COCO Karpathy test split~\citep{lin2014microsoft,karpathy2015deep} using CIDEr~\citep{vedantam2015cider} and CLIPScore~\citep{hessel2021clipscore}.

\paragraph{Multimodal understanding.}
We report POPE~\citep{li2023evaluating}, MMBench~\citep{liu2024mmbench}, SEED-Bench~\citep{li2024seed}, VQAv2~\citep{goyal2017making}, GQA~\citep{hudson2019gqa}, and OK-VQA~\citep{marino2019ok}. 

\paragraph{Image generation.}
GenEval~\citep{ghosh2023geneval} and DPG-Bench~\citep{hu2024ella} assess compositional generation and long-prompt adherence, respectively. Lower values are better for perplexity; higher values are better for the other reported metrics.

\subsection{Evaluation Sampling}
\label{app:evaluation-sampling}

Table~\ref{tab:appendix-sampling} gives the evaluation settings for visual question answering and text-to-image generation. For text outputs, the sampling steps apply to each generated chunk. The CFG sweep in Figure~\ref{fig:design-analysis}(b) varies the guidance scale instead of fixing it to the values below.

\begin{table}[H]
  \caption{Inference settings for visual question answering and image generation.}
  \label{tab:appendix-sampling}
  \centering
  \small
  \setlength{\tabcolsep}{6pt}
  \renewcommand{\arraystretch}{1.12}
  \begin{tabularx}{\linewidth}{@{}Xcccc@{}}
    \toprule
    Task & Sampler & Steps & CFG scale & SDE $\gamma$ \\
    \midrule
    Visual question answering & SDE & 16 & 3.0 & 1.0 \\
    Text-to-image generation & ODE & 64 & 5.0 & -- \\
    \bottomrule
  \end{tabularx}
\end{table}


\end{document}